\documentclass[runningheads]{llncs}
\usepackage[T1]{fontenc}
\usepackage{graphicx}
\usepackage{marvosym} 
\usepackage{color}
\usepackage{hyperref}
\usepackage{algorithm}
\usepackage{algorithmic}
\begin{document}
\title{Automatic Image Blending Algorithm Based on SAM and DINO} 
%
%
\author{Haochen Xue\inst{1*} \and
Mingyu Jin\inst{2*} \and
Chong Zhang\inst{1} \and
Yuxuan Huang\inst{1} \and
Qian Weng \inst{1} \and
Xiaobo Jin \inst{1}\textsuperscript{\Letter}}
\authorrunning{H. Xue, M. Jin, C. Zhang et al.}
%
\institute{School of Advanced Technology, Xi'an Jiaotong-Liverpool University, Suzhou, China \and
Electrical and Computer Engineering Northwestern University, Evanston, Illinois, USA
\\
 \email{u9o2n2@u.northwestern.edu} \\
\email{\{Haochen.Xue20, Chong.zhang19, Yuxuan.Huang2002, Qian.Weng22\}@student.xjtlu.edu.cn} \\
\email{Xiaobo.Jin@xjtlu.edu.cn}
\footnote{* Equal contribution. \\  \textsuperscript{\Letter} Corresponding author.}
}
\maketitle              
\begin{abstract}
The field of image blending has gained popularity in recent years for its ability to create visually stunning content. However, the current image blending algorithm has the following problems: 1) The manual creation of the image blending mask requires a lot of manpower and material resources; 2) The image blending algorithm cannot effectively solve the problems of brightness distortion and low resolution. To this end, we propose a new image blending method: it combines semantic object detection and segmentation with corresponding mask generation to automatically blend images, while a two-stage iterative algorithm based on our proposed new saturation loss and PAN algorithm to fix brightness distortion and low resolution issues. Results on publicly available datasets show that our method outperforms many classic image blending algorithms on various performance metrics such as PSNR \cite{karim2020measurement} and SSIM \cite{wang2004image}.

\end{abstract}

\section{Introduction}

Image blending is a versatile technique that can be used in a variety of applications where different images need to be combined to create a unified and visually appealing final image \cite{zhang2020deep} \cite{zhang2021image}. Essentially, it involves taking a selected part of an image (usually an object) and seamlessly integrating it into another image at a specified location. The ultimate goal of image fusion is to obtain a uniform and natural composite image. This task presents two significant challenges: (1) the accuracy of object cropping region is relatively low and (2) object cropping region may be inaccurate. Another challenge is that the blending process must adjust the appearance of cropped objects to match the new background.

GP-GAN and Poisson image editing are currently popular image blending methods \cite{wu2019gp}. In this method, to generate a high-resolution image, the user selects an object in the source image with an associated mask, and uses GP-GAN or Poisson to generate high-quality versions of the source and target images. However, the images generated by GP-GAN and Poisson Image Blending are not realistic. In addition, traditional algorithms often lead to brightness distortion problems in blended images, where composite images tend to exhibit excessive brightness in small clusters of pixels, compromising the overall realism of the image. To overcome this challenge, we reconstruct the blending algorithm for deep image blending \cite{zhang2020deep} using Pixel Aggregation Network (PAN) and a new loss function, iteratively improving the image blending process \cite{liu2018path,wang2019panet} . As a result, our blending algorithm produces images with consistent brightness, higher resolution, and smoother gradients.

\begin{figure}[t]
	\centering
	\includegraphics[width=1.0\columnwidth]{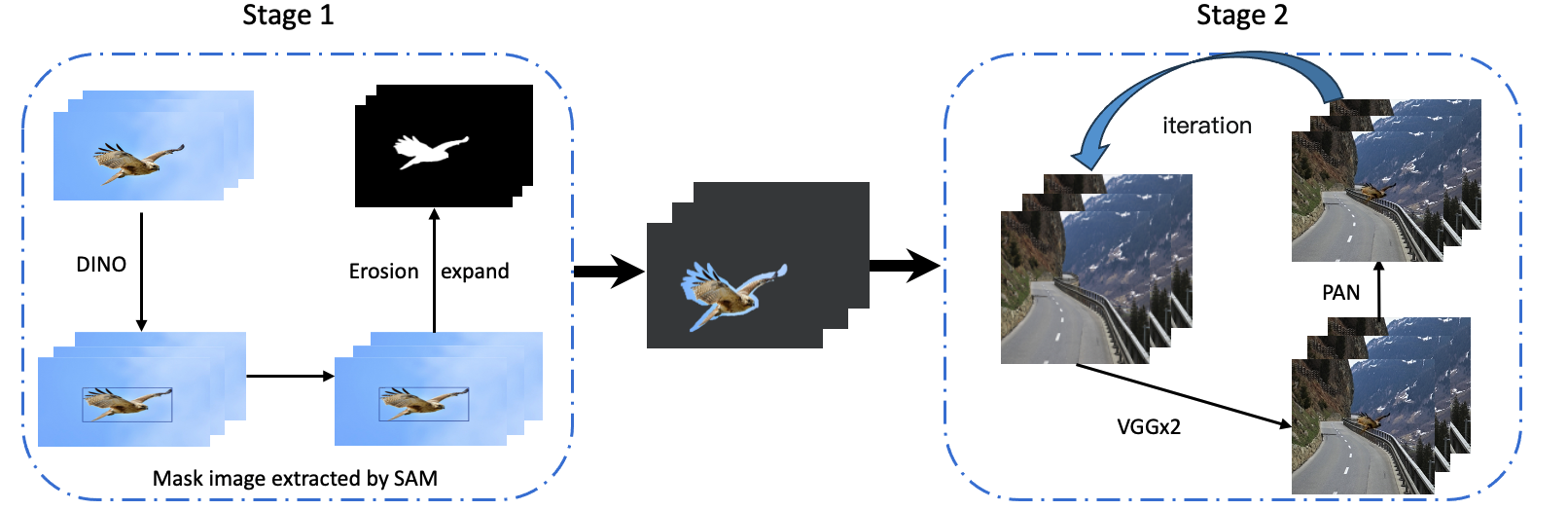} 
	\caption{Image blending refinement algorithm with automatic mask generation.}.
	\label{fig1-introduction}
	\vspace{-30pt}
\end{figure}

All image fusion algorithms require mask as input to crop the objects that need to be fused, but the mask images of the previous algorithms are all handmade, and these mask images are not accurate enough to represent the position of the foreground, which may lead to poor image fusion effect. The traditional segmentation method for automatically generating masks mainly includes RCNN \cite{girshick2014rich}, which was gradually replaced by more powerful methods, such as the SAM(Segment Anything) method proposed by Meta \cite{bharati2020deep}. However, SAM has its own limitations when it comes to image blending, as it tends to capture all objects \cite{xie2020sam} in a particular picture, whereas image blending only negatives one specific object in an image. To overcome this limitation, we apply DINO and use target text to distinguish our desired objects, resulting in better image blending. Our experiments show that the combination of DINO (DETR with Improved deNoising anchOr boxes) \cite{zhang2022dino} and SAM  generates more accurate masks than RCNN. But there remains a potential problem that other researchers may not have mentioned, namely that precisely segmenting objects may not always yield optimal results. If the mask image does not carry any relevant information of the original image, the blended image may lose important details \ref{fig1-introduction}. To address this challenge, we apply the classical erosion dilation step, which helps preserve important details in the original image for better blending results.

In our work, we try to solve the problem of low precision and low efficiency of manually cutting masks by generating masks through object detection and segmentation algorithms. These algorithms can make clipping operations easier and save time. In particular, we combine DINO and SAM algorithms to generate masks, DINO is used for target detection, and SAM can segment targets with the results of DINO, generating corresponding high-quality mask images. Because compared with the traditional RCNN algorithm, the mask of this algorithm can cover objects better and has stronger generalization ability. Aiming at the problem of sharp protrusions in the mask and the inability to carry source image information for fusion, we performed erosion and expansion operations on the mask, which will bring better image fusion results. Finally, due to the problems of low resolution and brightness distortion in the traditional hybrid algorithm, we also propose a new loss called saturation loss and further improve the effect of the algorithm through PAN. Evaluation metrics including PSNR, SSIM, and MSE on multiple image datasets show that our hybrid image can outperform previous hybrid models GP-GAN, Poisson Image, etc.

Our work has mainly contributed to the following aspects:

\begin{itemize}

\item We propose an automatic mask generation method based on object detection and SAM segmentation.

\item In image blending, we utilize erosion and dilation operations to manipulate the resulting mask for better image blending.

\item We propose a new loss function, called saturation loss, for use in deep image blending algorithms to address sudden contrast changes at the seams of blended images.

\item We use PAN to process blended images, solving problems associated with low image resolution and distortion of individual pixel gray values.

\end{itemize}

The overall paper is organized as follows: in Section 2, we introduce previous related work on image segmentation and detection as well as image blending; Section 3 gives a detailed introduction to our method; in Section 4, our algorithm will be compared with other algorithms. Subsequently, we summarize our algorithm and possible future research directions.

\section{Related Work}

\subsection{Image Blending}

The simplest approach to image blending (Copy-and-paste) is to directly copy pixels from the source image and paste them onto the destination image, but this technique can lead to noticeable artifacts due to sudden changes in intensity at the boundaries of the composition. Therefore, researchers have proposed various intelligent image blending algorithms to overcome this problem and produce more realistic composite images. These advanced methods use complex mathematical models to integrate source and destination images and improve the overall aesthetics of the blended image.

A traditional approach to image blending is Poisson image editing, first cited by Perez et al. \cite{DBLP:journals/tog/PerezGB03}, where it exploits the concept of solving Poisson's equation to blend images seamlessly and naturally. This method transforms the source and target images into the gradient domain, thus obtaining the gradient of the target image. Another image blending technique is the gradient domain blending algorithm, proposed by Perez et al. \cite{DBLP:conf/siggraph/LeventhalGS06}. The basic idea is to decompose the source image and the target image into the gradient domain and the Laplacian domain, and then use the weighted average to combine them. Deep Image Blending refers to the gradient in Poisson image editing, and making the gradient into a loss function, plus the loss of texture and content fidelity, Deep Image Blending produces higher image quality than Position image blending\cite{zhang2020deep}. Our work refers Deep Image Blending and optimized them to generate more realistic images.

Image inpainting is a technique in which a network uses learned semantic information and real image statistics to fill in missing pixels in an image \cite{ledig2017photo}. This process is done with deep learning models trained on large datasets, enabling them to learn the patterns and textures of images. With image inpainting, you can remove unwanted elements from an image or seamlessly restore damaged areas. The technique has many applications in areas such as image processing, computer graphics, and biomedical imaging. Besides image blending, there are several other popular image editing tools such as image denoising, image super-resolution, image inpainting, image harmonization, style transfer, etc. With the rise of generative adversarial networks (GANs), these editing tasks have seen significant improvements in the quality of generated results, e.g. GP-GAN \cite{goodfellow2020generative} \cite{arjovsky2017towards} \cite{alsaiari2019image}. Image super-resolution involves using deep learning models to learn image texture patterns and upsampling low-resolution images to high-resolution images. Our process will use PAN to achieve this target.

\subsection{Image Segmentation and Detection}

In the past, Regions with CNN features (RCNN) \cite{girshick2014rich} was the best region-based approach for semantic segmentation based on object detection. RCNN first extracts a large number of object proposals using selective search, and then computes CNN features for each object. It then classifies each region using class-specific linear SVMs \cite{cortes1995support}. RCNN can be used on top of various CNN structures and shows significant performance improvement over traditional CNN structures. But in our experiments, it did not perform as well as SAM. Because the CNN features in RCNN are not specially designed for image segmentation tasks, resulting in poor performance. Also, this feature does not provide enough spatial information for accurate boundary generation.

DINO is an advanced object detection and segmentation framework used by our pipeline to solve the problem of identifying the most important objects from segmented images via SAM \cite{chen2023semantic} . DINO introduces an improved anchor box and a mask prediction branch to implement a unified framework that can support all image segmentation tasks, including instance, bloom and semantic segmentation. The process involves the dot product of a high-resolution pixel embedding map with a query embedding generated by DINO to predict a set of binary masks that accurately identify the most important objects. Mask DINO is an extension of DINO that leverages this architecture to support more general image segmentation tasks. The model is trained end-to-end on a large-scale dataset and can accurately detect and segment objects in complex scenes. Mask DINO extends DINO's architecture and training process to support image segmentation tasks, making it an efficient tool for segmentation applications. This capability is especially useful for complex tasks, such as medical imaging and self-driving cars, where precise object recognition is critical for decision-making. Overall, the unified framework of DINO and Mask DINO provides a robust and accurate method for object detection and segmentation.

Essentially, an ideal image segmentation algorithm should be able to identify unknown or new objects by segmenting them from the rest of the image. Facebook Meta AI has developed a new advanced AI model called the "Segment Anything Model" (SAM)\cite{caron2021emerging} that can extract any object in an image with a single click. Segment Anything Model (SAM) leverages a combination of cutting-edge deep learning techniques and computer vision algorithms to accurately segment any object in an image in real time. SAM can efficiently cut out objects from any type of image, making the segmentation process faster and more precise. This new technology is a major breakthrough in the field of computer vision and image processing because it can save a lot of time and effort when editing images and videos. Whether it is a person, car, building, or any other object, SAM can accurately detect and segment objects with one click, making it easier to extract the desired elements from complex images. SAM utilizes contextual information to accurately segment objects in cluttered scenes, making it suitable for various real-world applications. The model is faster and more efficient than traditional segmentation algorithms, making it ideal for applications where real-time performance is critical, such as robotics, autonomous driving, and medical imaging. Additionally, one of SAM's unique features is its interactive segmentation capabilities. The model interactively segments objects in real time with just one click, allowing users to refine segmentation and easily explore different parts of the image. The SAM model uses advanced deep learning algorithms and computer vision techniques, enabling it to quickly analyze and understand complex images.

\section{Method}
In this section, we first give the overall framework of the algorithm, and then introduce the details of each algorithm one by one.

\begin{figure}[htp!]
	\centering
	\includegraphics[width=1.15\columnwidth]{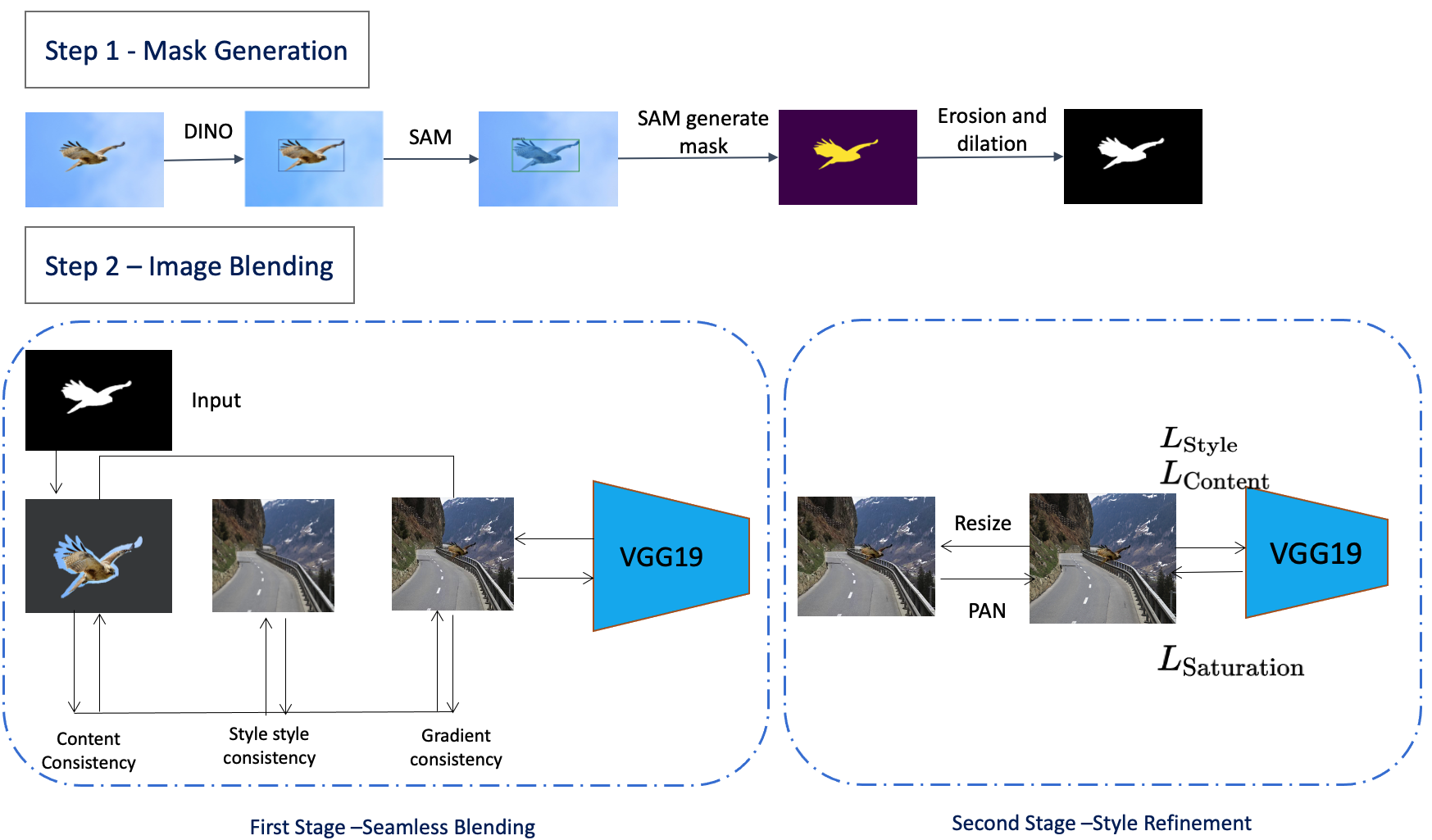}
	\caption{The method of automatic image blending} 
	\label{Method}
	\vspace{-10pt}
\end{figure}

\subsection{Framework of Our Method}
Our automatic image blending algorithm involves two stages. In the first stage, we use DINO to detect specific regions in an image based on textual descriptions and generate frames around objects, as shown in Figure \ref{fig3-SAM}. We then feed that frame into a SAM and extract the mask for that region. By combining DINO and SAM, we can accurately identify the target object in the image and generate an accurate mask, which saves time and effort compared to traditional methods. The resulting masks are subjected to erosion and dilation operations, and converted to black and white before image blending.

As shown in Fig. \ref{Method}, our two-stage image blending method solves the gradient problem of blended images, the style coordination problem, and preserves the blending details in the first stage. In the second stage, we further optimize the blended image using PAN to ensure that the resulting image has higher visual quality and fidelity, and that image details are not lost during iterations.

In the first stage of our algorithm, we update an input image with random values using three types of losses: gradient loss, content loss, and style loss. These losses help us obtain high-quality blended images. In the second stage of our algorithm, we further refine the blended image obtained from the first stage using "Pixel Adaptive Normalization" (PAN). We optimize the blended image for the content and style of the blended image and the target image. Furthermore, we incorporate our proposed "pixel mutation detection" to ensure that the generated images do not contain any artifacts or distortions.

\subsection{SAM and DINO}

We use DINO to detect specific regions in an image based on textual descriptions and generate a frame around the object, as shown in Figure \ref{fig3-SAM} (the input word is "bird"). We then feed the frame into a SAM and extract the mask for this region. By combining DINO and SAM, we solve the problem that SAM can only segment all objects instead of selecting specific objects. In Figure \ref{fig4-SAM}, we can observe that our algorithm can precisely identify the desired object in the image and generate an accurate mask. This method saves time and effort compared to traditional manual editing methods. It is worth noting that after getting the yellow-purple Mask, the mask image needs to be converted into black and white.

\begin{figure}[!htp]
	\centering
	\includegraphics[width=0.75\columnwidth]{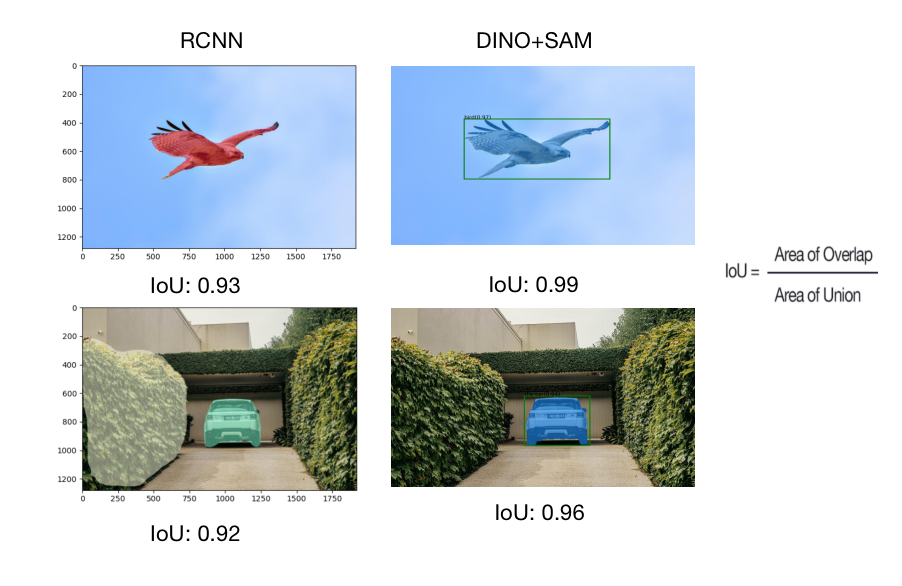}
	\caption{Comparison of results between traditional RCNN algorithm and DINO+SAM algorithm} 
	\label{IoU}
	\vspace{-10pt}
\end{figure}

\begin{figure}[!htp]
	\centering
	\includegraphics[width=0.75\columnwidth]{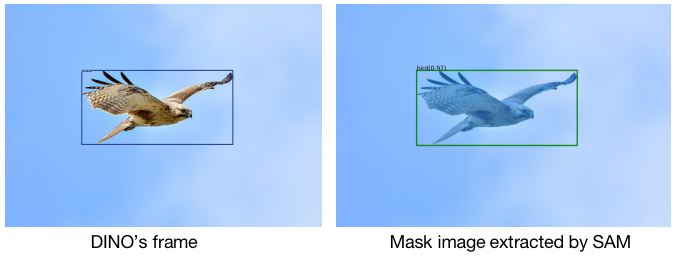} 
	\caption{Mask image extraction process by SAM}.
	\label{fig3-SAM}
	\vspace{-10pt}
\end{figure}

\begin{figure}[!htp]
	\centering
	\begin{minipage}[t]{0.36\textwidth}
		\centering
		\includegraphics[width=\textwidth]{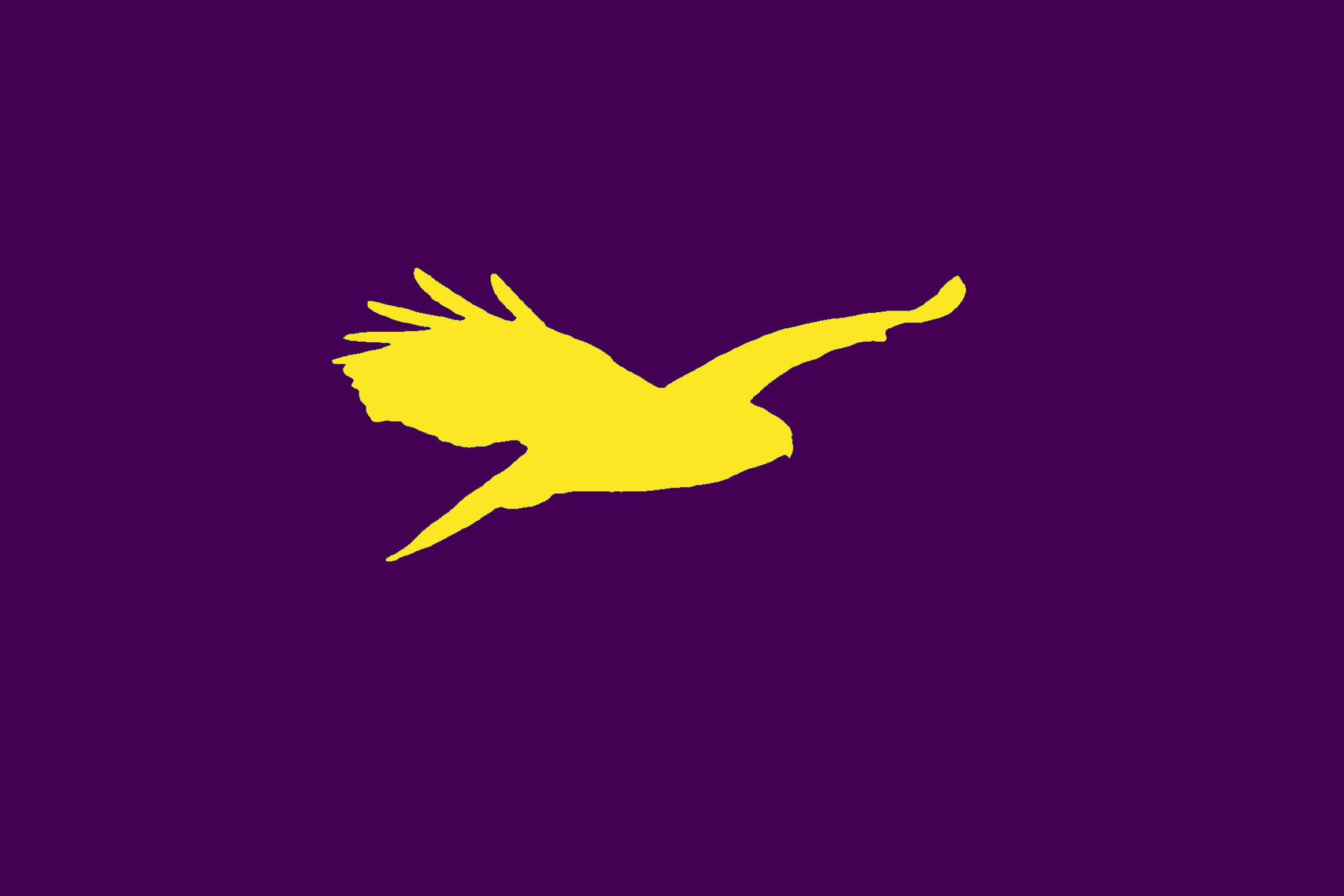}
	\end{minipage}
	\begin{minipage}[t]{0.36\textwidth}
		\centering
		\includegraphics[width=\textwidth]{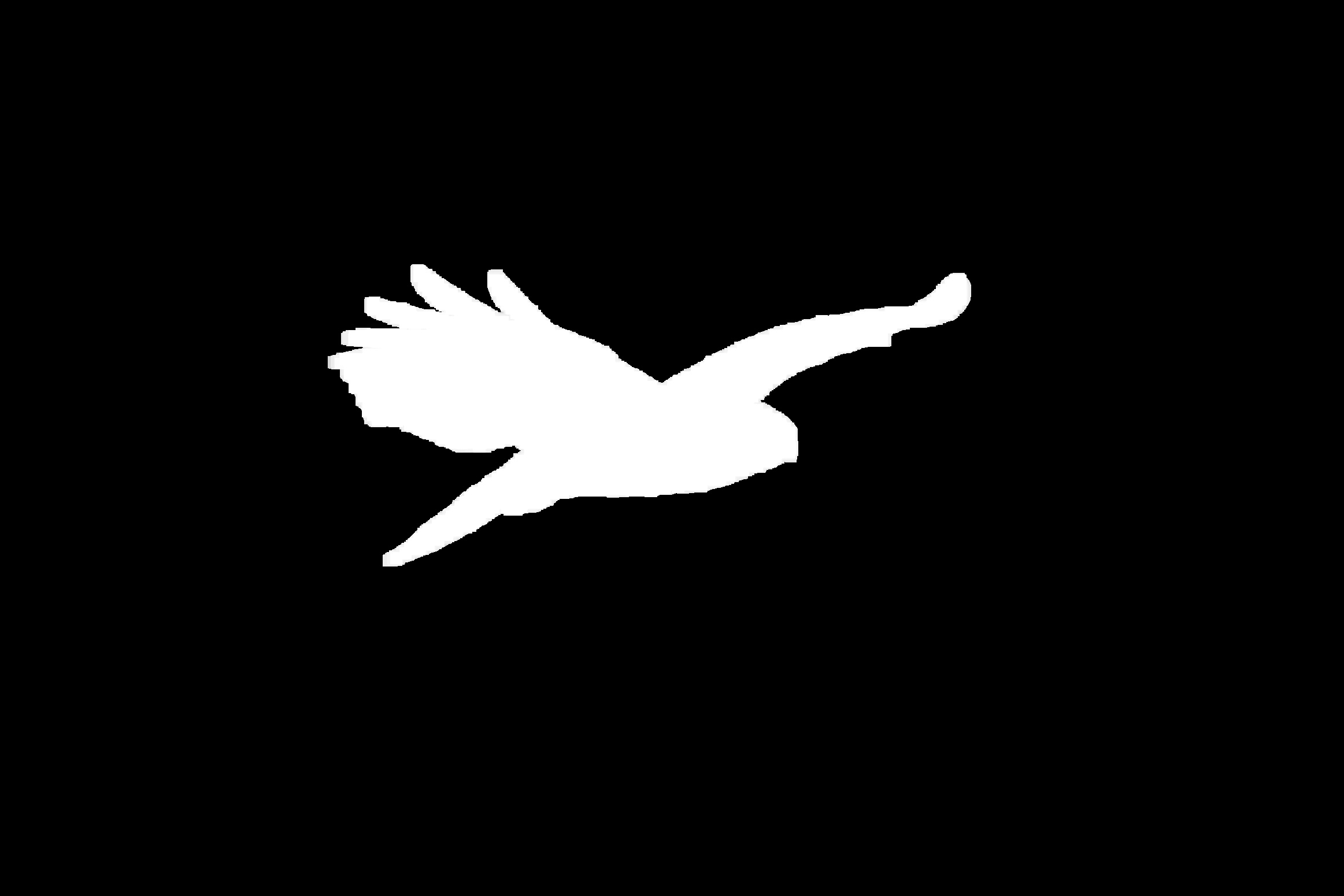}
	\end{minipage}
	\label{fig4-SAM}
	\vspace{-10pt}
 \caption{Mask image extracted by SAM: 1) left: before corrosion operation; 2) right: after corrosion operation}
\end{figure}

In terms of object detection, DINO has better performance than RCNN due to its self-supervised learning method and the advantages of Transformer network, which enables it to better capture global features. For semantic segmentation, SAM can better capture key information in images, and achieve more accurate pixel-level object segmentation through its multi-scale attention mechanism and attention to spatial features. Therefore, the mask generation achieved by the combination of these two methods is better than the traditional convolutional neural network, as shown in Fig. \ref{IoU}. We use IOU to measure the quality of the mask. It can be seen from the figure that the combination of DINO and SAM not only has a better segmentation effect visually, but also outperforms RCNN in terms of IOU.

\vspace{-10pt}
\subsection{Mask Refinement with Erosion and Dilation}

Erosion and expansion operations \cite{gonzalez2009digital} are performed on the mask to better refine the mask. The whole process of mask operation is as follows: First, an erosion operation is applied to shrink the sharp and misclassified edges of the mask. Secondly, perform a dilation operation on the eroded mask to expand its edges to ensure that the mask completely covers the target object and maintains a smooth boundary. By manipulating the mask in this way, we can improve the coverage of the mask, making it more suitable for image blending. In the image fusion stage, the processed mask can also carry part of $M_S$(source image) information, making the final fused image more natural.

\subsection{Two-Stage Deep Image Blending}
In the first stage, a style loss is used to transfer the style information of $M_B$ and $M_T$ to make the blended image more harmonious and realistic, where $M_B$ is the blended image and $M_T$ is the target image. Content loss is used to ensure the fidelity of the content in the $M_B$ image, and avoid the loss of details caused by content smearing caused by style migration. Gradient loss is used to smooth out blended edges. At this stage, through continuous iterations, the fusion edge will gradually become smoother, and the texture of the fusion object will gradually resemble $M_T$ without losing any details.

After the first stage of processing, the mixed edge of the object in the $M_B$ image is very smooth, but there are still significant differences between the mixed object and $M_T$ in terms of similarity and illumination, which may affect the quality and realism of the image. We need to continue optimizing images for this issue. The second stage takes the output image of the first stage as input, defined as $M_{br}$, where the style loss and content loss still exist, but the loss function will be calculated based on $M_{br}$. In addition to using the results produced by content loss and style loss optimization, we also propose a new saturation loss (detailed in 3.5) to calculate the pixel mutation between $M_{br}$ and $M_T$ to solve the mixed image The problem of unrealistic lighting and large contrast differences in the medium. Finally, the texture of the object in the generated result image is consistent with the source object $M_T$.

\subsection{Saturation Loss}

\begin{figure}
	\centering
	\includegraphics[width=0.83\columnwidth]{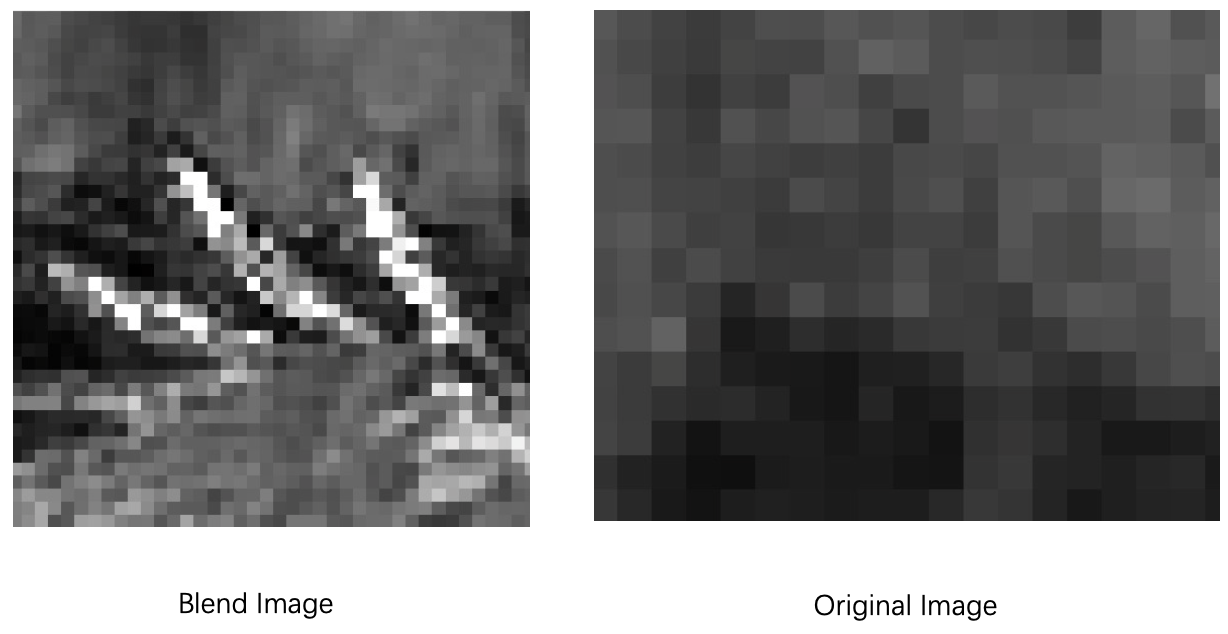} 
	\caption{Edge comparison between the fused image and the original image at the S-layer}.
	\label{loss1}
	\vspace{-20pt}
\end{figure}

Because the basic brightness of the blended background image and the target image is different, there will be a certain contrast difference after blending so that the naked eyes can perceive the existence of the blending operation. At different color coordinates, each coating of the fused image behaves differently. We observe that there are obvious differences at the fusion seams of R, G, and B layers under RGB color coordinates. However, after converting the fused image to the HSV color model, the Saturation layer of the fused image will have a sudden change in the saturation value at the edge where the source image and the target image are mixed, as shown in Fig. \ref{loss1}.
\vspace{-2pt}

\begin{figure}
	\centering
	\includegraphics[width=0.75\columnwidth]{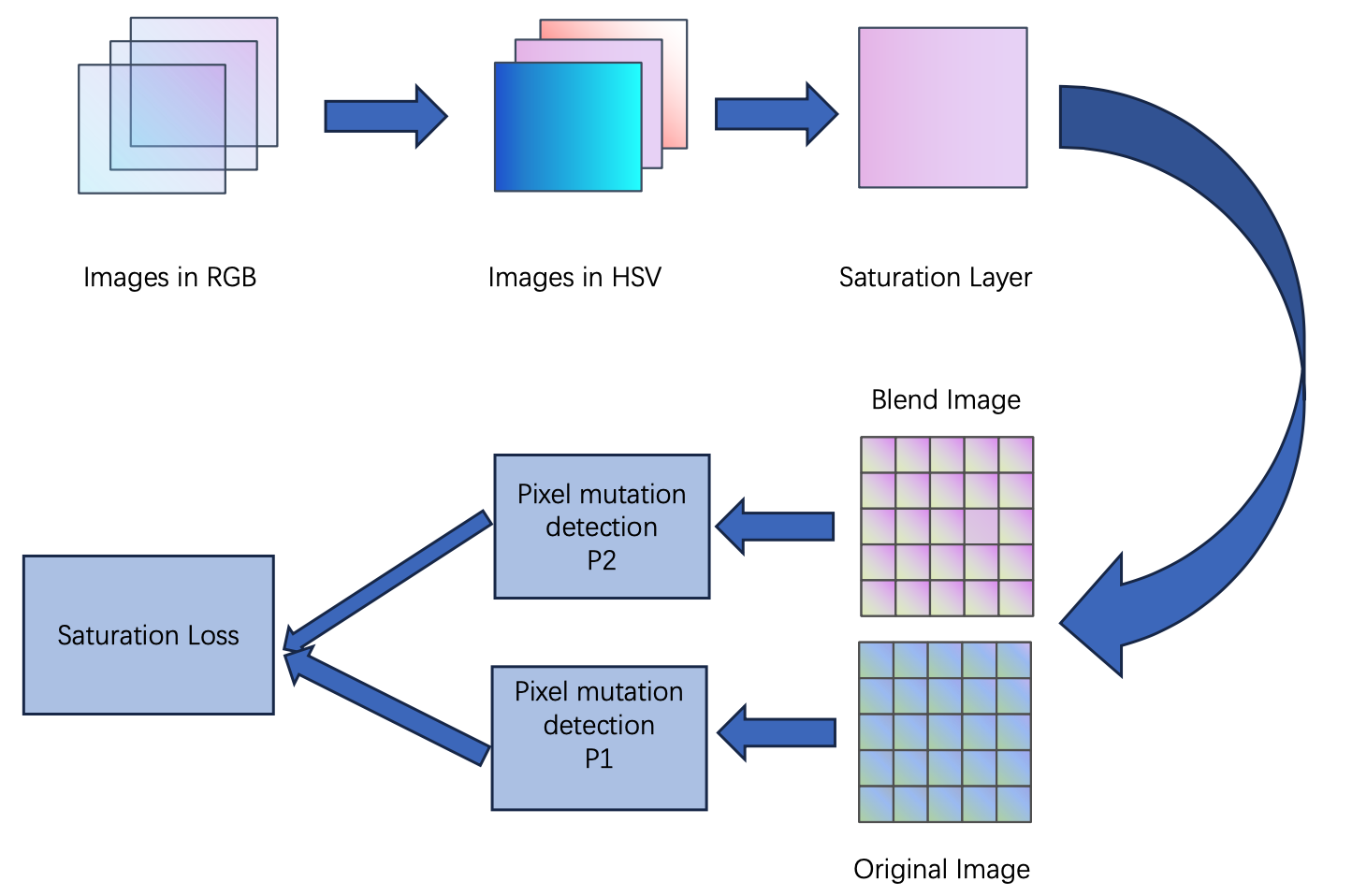} 
	\caption{Calculation framework of saturation loss.}
	\label{loss2}
	\vspace{-20pt}
\end{figure}

According to this characteristic of saturation, we propose a saturation loss to measure the authenticity of the fused image observed by the naked eye. The detailed process is shown in Fig. \ref{loss2}. First, we input the fused image set at the entrance, convert the RGB color coordinates to HSV color coordinates, and extract the Saturation layer of each fused image. The next step is to perform pixel mutation detection on the mixed image and the original image according to the following equation ($H$ and $W$ are the height and width of the image)
\begin{eqnarray}
		M & = & \sum_{m=1}^H \sum_{n=1}^W\left|P_{m+1, n}-P_{m, n}\right|+\left|P_{m, n+1}-P_{m, n}\right|, \nonumber\\
		L_{\mathrm{Sat}} & = & \frac{M_{\mathrm{ble}}-M_{\mathrm{ori}}}{H * W},
\end{eqnarray}
where $P(i,j)$ represents the saturation value of row $i$, column $j$ of the image, $M$ is the statistics of the amount of mutation in each S-layer image pixel, $M_{ble}$ and $M_{ori}$ is represents the statistics on the background image and the blended image. Finally, we calculate the saturation loss by the difference between the two sets of detection results.

\section{Experiments}

\subsection{Method Comparison}

We compared our algorithm with other algorithms qualitatively and quantitatively, and the results showed the superiority of our algorithm. The experimental settings shown in the below table \ref{tab:val-set}. In stage 1, the highest weight setting for gradient loss is because the main goal in the first stage is to solve the gradient problem and make the blending edges smoother. In the second stage, to address the lighting and texture issues of fused photos, style loss and saturation loss were designed to be $10^5$.
\vspace{-10pt}
\begin{table}[!htp]
\vspace{-10pt}
\caption{Experimental parameter settings}
\centering
\begin{tabular}{|l|l|l|l|l|}
\hline
Parameter  & Gradient weight & Style weight & Content weight & Saturation loss \\
\hline
Stage 1 & 1e4          & 1e3           & 1               & 0 \\   
Stage 2 & 0        & 1e5           & 1               & 1e5 \\       \hline
\end{tabular}
\vspace{-10pt}
\label{tab:val-set}

\end{table}
\vspace{-10pt}
\begin{figure}[htp!]
	\centering
	\includegraphics[width=1.1\columnwidth]{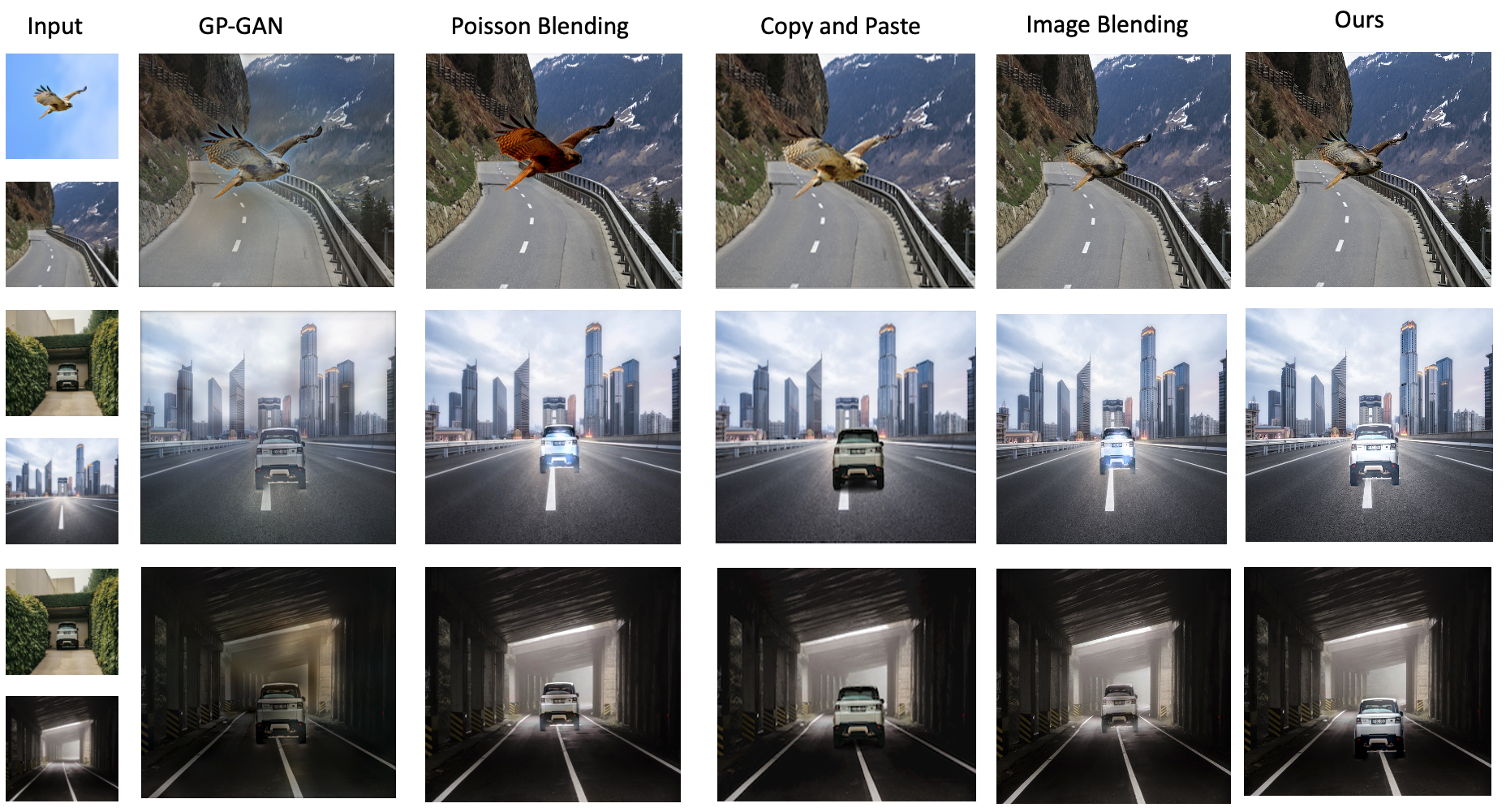} 
	\caption{Qualitative results comparison between our algorithm and GP-GAN, Poisson Blending, Copy and Paste approaches on multiple sets of images}
	\label{comparison}
 \vspace{-20pt}
\end{figure}

As shown in Fig. \ref{comparison}, the copy and paste method simply copies the source image to the corresponding location on the destination. Deep reconciliation requires training a neural network to learn visual patterns from a set of images and use them to create realistic compositions or remove unwanted elements. Poisson blending is another technique that seamlessly blends two images by considering their gradients and minimizing their differences. The technique involves solving Poisson's equation, which preserves the overall structure of images while describing the flow of color from one image to another. Finally, GP-GAN is a generative adversarial network (GAN) that uses a pretrained generator network to generate high-quality images similar to the training data. Generator networks are pretrained on large image datasets and then fine-tuned on smaller datasets to generate higher quality images. Unfortunately, these methods may not generalize well to our test cases and may result in unrealistic bounds and lighting. In the end, our algorithm produced the most visually appealing results for mixing borders, textures, and color lighting.

The copy-paste method of image fusion has some disadvantages. For example, the alignment of the target image to the background image needs to be precisely controlled, otherwise obvious incongruities and artifacts will result. Edges are treated to prevent jagged edges or visible machining marks. The problem of color uniformity is considered to prevent the fusion image style from being inconsistent. This makes the result have obvious artificial boundaries. Compared with these algorithms, we can clearly see that our two-stage algorithm adds more style and texture to the blended images. On the other hand, GP-GAN produces worse visual results in mixed boundary and color lighting. The overall color is dark and the edges are poorly processed. It brings rich colors to the raw edges of the image, resulting in inconsistent style and texture.

In order to quantitatively compare the performance of our and other methods, we compare them on various indicators including PSNR (peak signal-to-noise ratio), SSIM (structural similarity) and MSE (mean square error), and the results are shown in Table \ref{tab:exp_results}.  It can be seen that the average performance of our method achieves the best results on PNSR and SSIM, but MSE is slightly worse than Poisson Blending. This is mainly because our method does not simply migrate the source image to the target image, but further refines the mixed result with style and saturation consistency to make the generated picture more realistic, resulting in a slightly larger fitting error MSE. And compared with the image of Deep Image Blending, the images generated by our optimized model perform better in PSNR, SSIM, and MSE, which also indicates that the superiority of our model.  

\vspace{-10pt}
\begin{table}[!htp]
\caption{Quantitative comparison of average results between our method and other methods on PSNR, SSIM and MSE metrics}
	\centering
	\begin{tabular}{|l|c|c|c|}
		\hline
		Method & PNSR & SSIM & MSE           \\ 
  \hline
		GP-GAN & 18.94  & 0.74           & 829.07    \\ 
		Poisson Blending & 21.38 & 0.70 & \textbf{473.45} \\ 
            Deep Image Blending & 22.03 & 0.73 & 723.71\\
		Copy and Paste & 18.89 & 0.56 & 839.15 \\
        Ours & \textbf{23.07} & \textbf{0.77} & 695.60 \\
        \hline
	\end{tabular}
	\label{tab:exp_results}
	\setlength{\abovecaptionskip}{2pt}
\end{table}


\subsection{Ablation Study}
\begin{figure}[!htp]
    \vspace{-30pt}
	\centering
	\includegraphics[width=0.95\columnwidth]{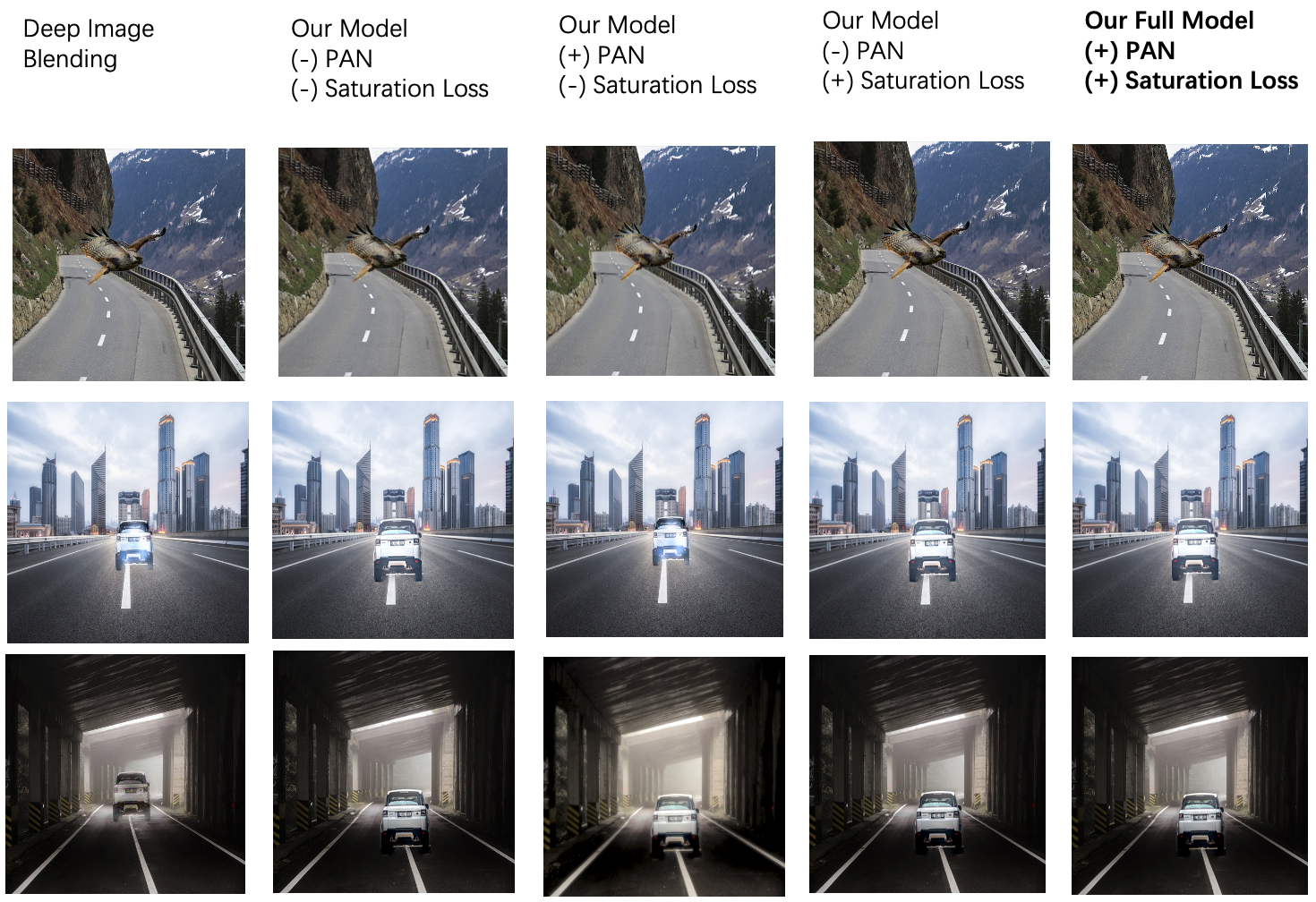} 
	\caption{The results of the ablation experiment: (+) and (-) respectively indicate that a certain part of the algorithm participates or does not participate.}.
	\label{abl-res}
    \vspace{-30pt}
\end{figure}

Let's take three pictures as an example to conduct ablation experiments to analyze the role of PAN component and saturation loss in our method.

\textbf{PAN in Blending Refinement} We will only keep the saturation loss and remove PAN component from the model. The reduction in image resolution is substantial, resulting in a significant loss of image clarity.

\textbf{Saturation loss in Deep Image Blending} In this experiment, we will only keep PAN component and remove the saturation loss from the model.

Figure 6 shows the visualization results of 4 ablation experiments, which show that when using two components at the same time, the generated results are more realistic, and the style of the source image and the target image are more consistent, especially the blending margin area.

\begin{table}[!htp]
\vspace{-15pt}
\caption{Quantitative results of our algorithm's ablation experiments on 3 sets of images, where the three values in each cell represent the results on different image.}
	\begin{tabular}{|l|c|c|c|}
		\hline
		Metrics & PSNR                       & SSIM                    & MSE                           \\ \hline
		+PAN                           & 20.77/20.16/22.09          & 0.74/0.69/0.79          & 543.98/721.69/401.49          \\ 
		+Saturation Loss               & 19.79/19.94/22.29          & 0.6/0.62/0.81           & 681.91/658.20/383.38          \\ 
		\textbf{+PAN+Saturation Loss} & \textbf{29.57/24.58/23.64} & \textbf{0.72/0.83/0.79} & \textbf{612.95/568.47/402.93} \\ 
		Deep Image Blending                       & 17.85/18.32/17.99          & 0.57/0.61/0.67          & 718.20/661.94/594.32          \\ \hline
	\end{tabular}
	\label{lab: ab-res}
 \vspace{-15pt}
\end{table}

Table \ref{lab: ab-res} further gives the quantitative results of the ablation experiments. The three numbers in each grid in the presentation represent the experimental results on 3 images. In terms of the first metric PSNR, our model outperforms the other two methods. Regarding SSIM, our model shows better results than methods lacking saturation loss, and performs similarly to methods lacking PAN. In terms of MSE, the performance of our model is mediocre relative to the other two methods. These metrics reflect the quality of generated images to some extent.
\vspace{-5pt}
\section{Conclusion}
\vspace{-5pt}
In our work, we address the low accuracy and low efficiency of manually cutting masks by generating masks through object detection and segmentation algorithms. Specifically, we combine DINO and SAM algorithms to generate masks. Compared with the traditional RCNN algorithm, the mask of this algorithm can cover objects better and has stronger generalization ability. We perform erosion and dilation operations on the mask to avoid sharp protrusions in the mask. Finally, we also propose a new loss, called the saturation loss, to address brightness distortion in generated images. Results on multiple image datasets show that our method can outperform previous image fusion methods GP-GAN, Poisson Image, etc.
\vspace{-5pt}
\section{Future Work}
\vspace{-2pt}
Future work includes proposing new evaluation criteria to better reflect human perception and aesthetics to improve the objectivity and accuracy of the model. Another potential research direction is how to deal with object occlusion in image fusion.

\bibliographystyle{plain}
\bibliography{main}

\end{document}